\documentclass[journal]{IEEEtran}

\IEEEoverridecommandlockouts                              

\usepackage{numbersUncertainty}

\newcommand{\cnn}{\textsc{cnn}\xspace}

\newcommand{\kf}{\textsc{kf}\xspace}
\newcommand{\ssd}{\textsc{ssd}\xspace}
\newcommand{\sort}{\textsc{sort}\xspace}
\newcommand{\nms}{\textsc{nms}\xspace}

\newcommand{\rcnn}{\textsc{r-cnn}\xspace}
\newcommand{\yolo}{\textsc{yolo}\xspace}
\newcommand{\yolox}{\textsc{yolo-x}\xspace}

\title{ Extended target tracking utilizing machine-learning software -- with applications to animal classification \\
}

\author{Magnus Malmstr{\"o}m, Anton Kullberg, Isaac Skog \IEEEmembership{Senior Member, IEEE}, Daniel Axehill \IEEEmembership{Senior Member, IEEE} , Fredrik Gustafsson \IEEEmembership{Fellow, IEEE} 
	\thanks{This work is supported by Sweden's innovation agency, Vinnova, through project iQDeep (project number 2018-02700). The authors would like to express their gratitude for the image data provided by Norwegian institute for nature research (NINA), and the Large Carnivore Center (Rovdjurscentret De 5 Stora), with financial support from WWF, for access to real-time camera trap images.  }
	\thanks{Magnus Malmstr{\"o}m, Anton Kullberg, Daniel Axehill, and Fredrik Gustafsson, are with Link{\"o}ping University (e-mail: \{magnus.malmstrom, anton.kullberg, daniel.axehill fredrik.gustafsson\}@liu.se)  }
\thanks{Isaac Skog, is with Uppsala University  (e-mail:  isaac.skog@angstrom.uu.se)}
}

\begin{document}
\maketitle
 \bstctlcite{IEEEexample:BSTcontrol}

\thispagestyle{empty}
\pagestyle{empty}


\begin{abstract}
    This paper considers the problem of detecting and tracking objects in a sequence of images.
    The problem is formulated in a filtering framework, using the output of object-detection algorithms as measurements.
    An extension to the filtering formulation is proposed that incorporates class information from the previous frame to robustify the classification, even if the object-detection algorithm outputs an incorrect prediction.
    Further, the properties of the object-detection algorithm are exploited to quantify the uncertainty of the bounding box detection in each frame.
    The complete filtering method is evaluated on camera trap images of the four large Swedish carnivores, bear, lynx, wolf, and wolverine. 
    The experiments show that the class tracking formulation leads to a more robust classification.
\end{abstract}


\section{Introduction}
This paper considers the problem of detecting and classifying objects in a sequence of images or a video and tracking them over time. In particular, it investigates how to incorporate information of the object's class to improve the robustness of the tracking algorithm.
As object detection through neural networks (\nn{s}) becomes widely used in safety-critical applications such as self-driving cars, it becomes of great importance that their predictions are robust and trustworthy. For example, if a pedestrian crosses the road, the car should detect the pedestrian in time to brake to avoid a collision. 
Further, it is also important to distinguish between different classes of objects, as this may influence the subsequent decision process.

Presented with a sequence of images, it is likely that the detected object belongs to the same class for the entire sequence.
By classifying many images assumed to belong to the same class, the probability of correct classification has been shown to increase \cite{braca2022statistical}.
Hence, even though there is an error in a particular \nn classification, it should be possible to correct the mistake by using information from classifications of previous images in the sequence. 
The problem can be split into two steps. Firstly, locate the object and classify it using an object-detection algorithm. Secondly, track the object over time, e.g., using a filter.

Lately, there have been numerous algorithms developed to solve the object-detection problem, e.g., Single Shot MultiBox detector (\ssd) \cite{liu2016ssd}, you only look once (\yolo) \cite{redmon2016yolo} and its extension \cite{ge2021yolox}, region-based convolutional \nn{s} (\rcnn) \cite{girshick2014rich} and its extension \cite{girshick2015fast,ren2015faster,chen2020relation}, and CenterNet \cite{duan2019centernet,zhou2019objects}. These algorithms find and classify the object in the image, while none follow it over time. 

There has been substantial work on developing algorithms that track bounding boxes in images, e.g.,  \cite{danelljan2019atom,bertinetto2016fully,marvasti2021deep,zhang2023learning,wang2018robust,zhang2022bytetrack,bewley2016simple,wang2020towards,han2018advanced}.
They usually consider one out of two problem formulations that use different solution strategies. The first problem formulation is called visual object tracking, where an object should be tracked over time given a reference frame. This problem is solved by introducing new \nn architectures where the information from the reference is incorporated \cite{danelljan2019atom,bertinetto2016fully,marvasti2021deep,zhang2023learning,wang2018robust}.
The second problem is called multi-object tracking, where, given a video sequence, all objects of interest should be tracked over time. The standard strategy to solve these problems is to view the output of standard object detection algorithms as a measurement for a filter used in a standard target tracking formulation \cite{zhang2022bytetrack,bewley2016simple,wang2020towards}. 
Two examples of algorithms that aim to solve multi-object tracking are ByteTrack \cite{zhang2022bytetrack} and \sort \cite{bewley2016simple}, which both rely on Kalman filters (\kf) and simple motion models to track the object. As a detection algorithm,  ByteTrack uses \yolox  \cite{ge2021yolox} and \sort uses Faster-\rcnn \cite{ren2015faster}, where here \ssd is used. 
However, neither of the methods track the object's class, which is of interest in safety-critical applications where some classes might be of higher importance than others. Nor do they specify the uncertainty in the measurement of the bounding boxes to be tracked. 
Previous work has included class information in a tracking framework to make the association step more robust \cite{gaglione2020classification}. 
Thus, accurately tracking the class of the object(s) in the scene is of high interest.

The contribution of this work is threefold. 
Firstly, we formulate the tracking and detection of an object in a sequence of images as a filtering problem, where the measurements come from a standard object detection algorithm.
The standard problem formulation is extended such that the uncertainty in the position of the bounding boxes is estimated. Secondly, we propose a method to  systematically adjust how much information regarding the object's class from previous frames should be considered. This paper shows that including the class information from previous frames improves the robustness of the tracking algorithm when considering lost tracks. 
Thirdly, the method is evaluated on a challenging task using camera trap images collected in Swedish forests for an animal conservation project.

\section{Extended object tracking} \label{sec:tracking}
Consider the problem of tracking an object and its class in a sequence of images given detections from a detection algorithm such as \ssd. It will be assumed that every image only consists of one object to make the notation more concise. However, the method can easily be extended to cover several objects using an association process based on the intersection over union (IoU) between the bounding boxes. 

\subsection{States in the tracking algorithm}
Denote $x \in \mathbb{R}^{n_x}$ as an image with $n_x$ pixels, i.e., the input data to the detection algorithm, and $y_n \in \{1, \ldots, M\! + \! 1 \}$ denotes the $M$ class labels of the object in the image and the background class.
Define the states 
\begin{subequations}
    \begin{align} \label{eq:state_box} %
    \chi^b  &= \begin{bmatrix} p_x & p_y & l & h \end{bmatrix}^\top,  \text{ and } \\
    [\chi^c]_m &= p(y=m|x), \quad m= 1,\ldots, M+  1, \label{eq:state_box_class}
    \end{align}
\end{subequations}
where $\chi^b \in \mathbb{R}^4$ represents the position and size of the bounding box of an object in the image and $\chi^c\in \mathbb{R}^{M+1}$  the confidence in the different classes and the background class in that box. Here $[\cdot]_m$ denotes the $m$'th element of a vector, i.e., in \eqref{eq:state_box_class}, it is used to denote the probability that the object belongs to the $m$'th class. 
Further, $(p_x, p_y)$ is the center, and $l$ and $h$ are the length and height of the bounding box, respectively.

At each time $t$, assume that a measurement $z^b_t=\chi^b_t + \bm{e}_t$ of the bounding box position and its size is available, with measurement noise $\bm{e}_t\sim \mathcal{N}(0, R^b_t)$. Here, $R^b_t$ is the covariance of the estimated bounding boxes from the object detection algorithm.
The position and size of the bounding box are assumed to follow a linear motion model with additive process noise $\bm{v}_t \sim \mathcal{N}(0, Q)$.
The covariance of the process noise $Q$ could be class dependent \cite{soldi2021space}, e.g., different classes move at different speeds.
Since the state-space model is linear, a \kf can be used to solve the filtering problem. In the experiment, a constant position motion model is assumed.  

\subsection{Robust classification} 
Assume that the object's class in the image is categorically distributed and that the state $\chi^c_t$ stays the same between the images, i.e., $\chi^c_t = \chi^c_{t-1}$.  
An estimate of the probability vector for the categorical distribution is given by $\hat{\chi}^c_t$. A measurement of the probability for the object's class is given by  $z^c_t$. 
Under the assumption that the estimate of the probability at time $t-1$ influences the estimated probability at time $t$, i.e., the same object is tracked over time, this influence should be included in the measurement update. Using a filtering formulation, this results in     
\begin{align} \label{eq:estimated_class}
\hat{\chi}_{t}^c &= (1-K^c_t) \hat{\chi}_{t-1}^c + K^c_t z^c_t
\end{align}
where $K^c_t\in [0,1]$ weighs how much impact the measurement of the class probability at time $t$ should have on the estimated \pmf for the object's class in the image sequence.
The formulation in \eqref{eq:estimated_class} makes the tracking algorithm more robust against ``incorrect'' measurements, where $1-K^c_t$ can be interpreted as a forgetting factor of the object's class.
There are many different approaches to selecting $K^c_t$, e.g., formulating an optimization problem to weigh the influence between measurements and old states, using a forgetting factor or the median.
In this paper, the value of $K^c_t$ will be selected such that the estimated state $\hat{\chi}_{t}^c$ is an average of the previous measurements and the prior, i.e., $K^c_t= 1/(t+1)$.
This choice is reasonable since if an object has been seen for a long time, it is unlikely its class would change, i.e., when $t \to \infty$, then $K^c_t \to 0$.

\begin{figure}[tb!]
	\centering
\scalebox{0.4}{
  \begin{tikzpicture}[scale=0.5,node distance=2cm]
    \node[inner sep=25pt] (start) at (0,0)
    {\includegraphics[width=.1\textwidth,trim={0cm 0cm 17cm 0cm}]{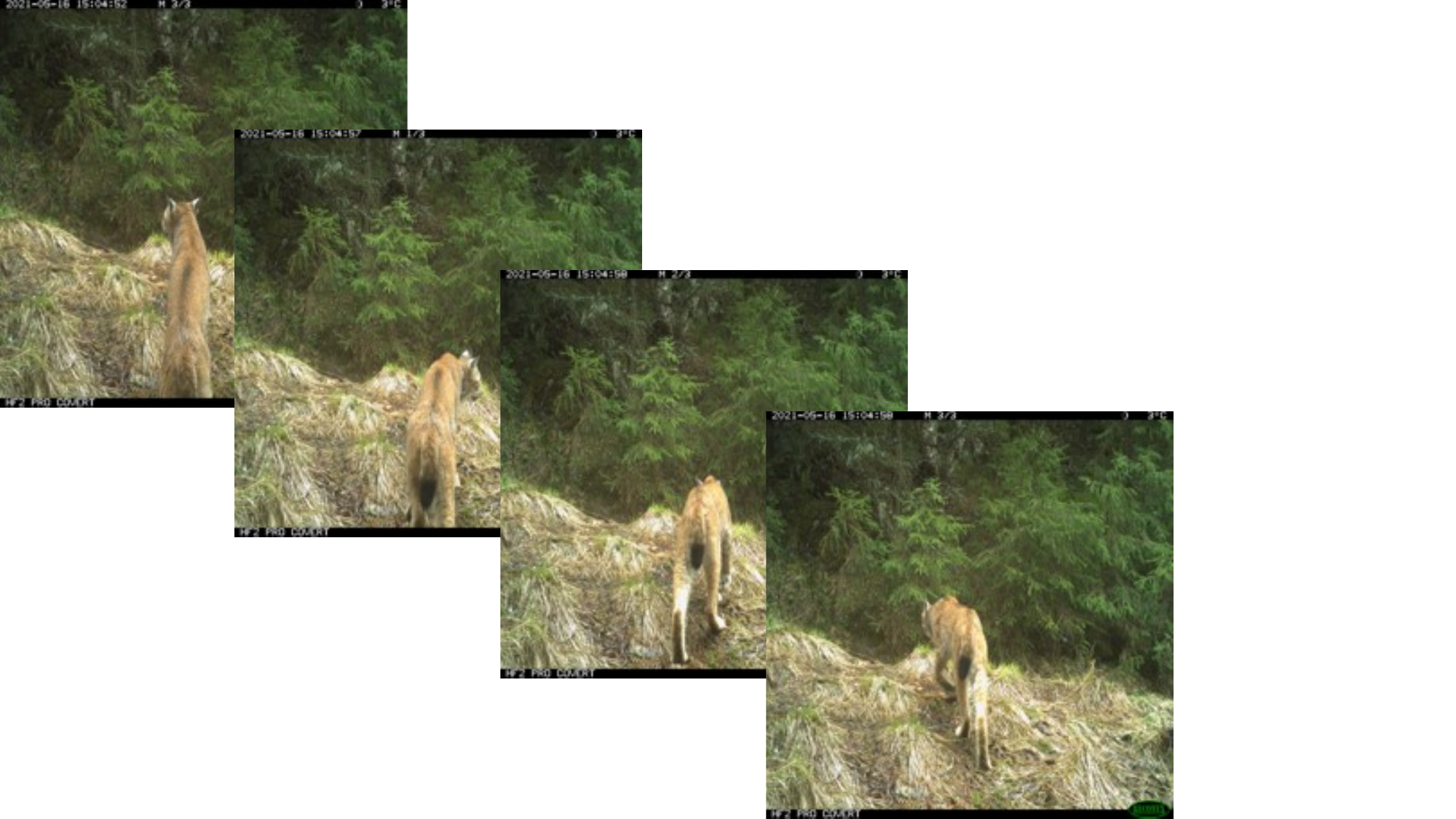}};
\node (object) [process, right of=start,xshift=2.5cm] {Measurement model, e.g., \ssd}; 
\node (filter) [process, right of=object,xshift=4cm] {Filtering algorithm, e.g., \kf \textcolor{blue}{and \eqref{eq:estimated_class}}};

\node (robust) [process, right of=filter, xshift=4cm] { Robust classification of tracks};

\draw [arrow] (start.east) -- node[anchor=north] {$x_t$} (object);
\draw [arrow] (object) --node[anchor=north] {\textcolor{blue}{$z^b_t,R_t^b,z_t^c$}}  node[anchor=south] {\textcolor{red}{$z_t^{b, (\ast)}, z_t^{c, (\ast)}$}} (filter);
\draw [arrow] (filter) -- node[anchor=north] {\textcolor{blue}{$\hat{\chi}_t^b, \hat{\chi}_t^c$}} node[anchor=south] {\textcolor{red}{$ \hat{\chi}_t^b,  z_t^{c, (\ast)} $}} (robust);
\coordinate[below of = filter, name = empty] {};
\draw (robust)  |-  (empty);
\draw [arrow] (empty)  -- node[anchor=west] {\textcolor{blue}{$\hat{\chi}_{t-1}^b,\hat{\chi}_{t-1}^c$}} node[anchor=east] {\textcolor{red}{$\hat{\chi}_{t-1}^b$}}  (filter);
\coordinate[right of = robust, name = empty_out] {};
\draw [arrow] (robust) -- (empty_out);


\end{tikzpicture}

  }
\caption{ Schematic illustration of the suggested filtering framework. In red are the quantities commonly used in multi-object tracking applications, and in blue are the quantities used in this paper. Two differences should be highlighted. Firstly, the measurement updates for the class probabilities $\hat{\chi}^c_t$ and secondly, the use of multiple anchor boxes such that the covariance of the estimated positions $R^b_t$ can be included in the \kf, where traditionally only the information of the most probable bounding box $z^{b,(\ast)}_t$ is used, see \eqref{eq:mostprobable} and \eqref{eq:proposals}.  }
  \label{fig:block_diagram}
\end{figure}
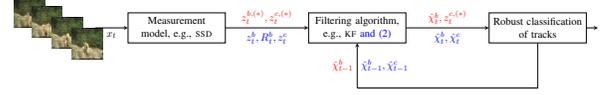

A schematic illustration of the filtering framework to solve the tracking problem can be seen in \cref{fig:block_diagram}. Here, the filtering algorithm includes information of the object's class using \eqref{eq:estimated_class}. Note that other methods, e.g., ByteTrack and \sort, can also easily be extended to include the information of the object's class by extending the used filtering algorithm with, e.g., \eqref{eq:estimated_class}.

This paper focuses on making the filter formulation more robust toward incorrect classifications.
In a target tracking framework, a {\em track} is often defined as the estimated history of a target.
In such a framework, it is crucial to know when an object appears or disappears from the sensor's field of view to kill and give birth to new tracks.
Here, the estimated probability mass function (\pmf) $\hat{\chi}_{t}^c$ is used as a surrogate to determine whether to kill a track, e.g., if $\max \hat{\chi}_{t}^c$ is below a given threshold the track is killed. Similarly, a new track can be born if $\max \hat{\chi}_{t}^c$ is above some threshold.

\section{Measurement model}
This section will cover how to use standard object-detection algorithms to generate measurements for a tracking algorithm. 
\subsection{Object detection}
Consider the problem of learning a detector used to detect and classify objects in an image for the dataset 
\begin{align}
\mathcal{T} \triangleq\{x_n,y_n^j,b_n^j\}_{n=1}^N, \quad j = 1, \ldots, J_n.
\end{align}
Here $y_n^j \in \{1, \ldots, M \}$ is the class label of the object, and $b_n^j \in \mathbb{R}^{4}$ is the shape of the bounding box in which the object is located. The subindex $j$ denotes the $j$'th object of the $J_n$ objects in the image. 
From a statistical point of view, learning the detector can be formulated as a system identification problem where one simultaneously identifies a model $f^b(x;\theta)$ for the bounding boxes and a model $f^c(x;\theta)$ for the conditional \pmf  $ p(y = m |x), \quad m=1, \ldots, M+1$ of a categorical distribution, where an extra class for the background is added. 
Here $\theta \in \mathbb{R}^{n_{\theta}}$ denotes the $n_{\theta}$ dimensional parameter vector of the parametrized model. 

The models $f^b(x;\theta)$ and $f^c(x;\theta)$ are often based on a pre-trained convolutional \nn (\cnn) \cite{liu2016ssd,redmon2016yolo,ge2021yolox,girshick2014rich,girshick2015fast,ren2015faster}, here referred to as the backbone \nn. 
The parameters in the model are the weights and biases of the \cnn. The superscript $c$ stands for classification and $b$ for bounding box.
\subsection{Single Shot MultiBox detector}
This paper focuses on using \ssd \cite{liu2016ssd} as the detection algorithm. However, the proposed method is more general and could be applied to other detection algorithms that use anchor boxes, e.g., \yolo. Here, anchor boxes are predefined boxes bounding the object in the images, where the boxes slide over the image. One of the key contributions of this paper is how to use the anchor boxes to compute the measurements in the tracking algorithm such that the covariance of the measurements is included, which is not common in the literature. The knowledge of the covariance simplifies the tuning of the \kf.

For \ssd, the backbone \nn is branched off at $R$ different hidden layers, where each branch is responsible for detecting objects of different sizes. The classification and bounding box regression will be split into different branches.
Each of those branches represents a predetermined grid. For grid $r$ with $\gamma_r$ grid points, $\alpha_r$ predetermined anchor boxes are specified. Then, anchor boxes are placed at every grid point.  
That is, for each image, the \ssd detects $N_b=\prod_{r=1}^{R} \gamma_r \alpha_r$ bounding boxes with corresponding confidence per class, i.e., $ f^c(x;\theta)^{(i)}$, and $f^b(x;\theta)^{(i)}$ where $i = 1\ldots N_b$.
%

%
The estimate of the model parameters is given by 
\begin{subequations}
    \begin{align}
    \hat{\theta}_N & = \argmax_{\theta} L_N(\theta), \\
    L_N(\theta) &= \sum_{n=1}^{N} \frac{1}{N_m}\big (L_c(\theta,x_n,y_n)  +  \alpha L_b(\theta,x_n,y_n,b_n)\big) 
    \end{align} 
\end{subequations}
where $L_N(\theta)$ is the loss function, which is the weighted sum between a classification loss $L_c$ and a location loss $L_b$, using the weighting parameter $\alpha$ \cite{liu2016ssd}. Here, $N_m$ is the number of matched boxes, i.e., boxes with a confidence of a non-background class larger than some predetermined threshold. Define the so-called positive indicator variables $\xi_{ij}^{y^j}= \{0,1 \}$ and negative indicator variable $\xi_{i}^{-}= \{0,1 \}$. The positive indicator variable is equal to one if the predicted bounding box $i$ matches the ground-truth bounding box $j$ with the class label $y^j$, and the negative indicator variable is used to indicate that the predicted bounding box does not overlap with any of the ground-truth bounding boxes. The classification loss is based on the assumption that the classes (including the background class) in the boxes are categorically distributed. Hence, the classification loss is given as  
\begin{subequations}
    \begin{align}
    L_c(\theta,x_n,y_n)  = & - \sum_{i=1}^{N_b} \sum_{j=1}^{J_n} \xi_{ij}^{y^j_n} \log(f_{y^j_n}^c(x_n; \theta)^{(i)}) \nonumber \\
    &  - \sum_{i=1}^{N_b} \xi_{i}^{-} \log(f_{M+1}^c(x_n;\theta)^{(i)}),
    \end{align}
    where both boxes containing objects and boxes not containing objects are represented. 
    The localization loss was chosen such that
    \begin{align}
    L_b(\theta,x_n,y_n,b_n) = \sum_{i = 1}^{N_b} \sum_{j=1}^{J_n} \xi_{ij}^{y^j_n} l_{L1}(f^b(x;\hat{\theta}_N)^{(i)}- b_n^j)
    \end{align}
    where $l_{L1}$ is the so-called smooth $L1$ loss defined as
	$l_{L1}(x) \triangleq \{|| x ||_2^2 / (2 \xi), \text{ } ||x ||_1< \xi, \text{ } || x ||_1 - 0.5 \xi, \text{ } \text{otherwise}\}$. 
\end{subequations}
Here $||.||_i$ is used to define the $i$'th norm of the vector.

In the prediction phase, non-maximum suppression (\nms) is typically used to remove overlapping boxes and boxes with too low confidence of the most probable class (excluding the background class), hence only keeping one box per object in the image. 
Define the most probable class as 
\begin{align}\label{eq:mostprobable}
\hat{y}^{\ast} & = \argmax_{m=1,\ldots, M} f^c_m(x;\hat{\theta}_N)^{(\ast)},
\end{align}
where $\ast$ indicates the index of the bounding box with the highest confidence of including an object of a non-background class of the $N_b$ predicted boxes, if there are multiple boxes with high confidence of an object for which there is no overlap, they are stored as separate objects.

\begin{figure}[tb!]
	\centering
	\includegraphics[trim={1cm 6cm 0.5cm 1cm},clip,width=0.355\textwidth]{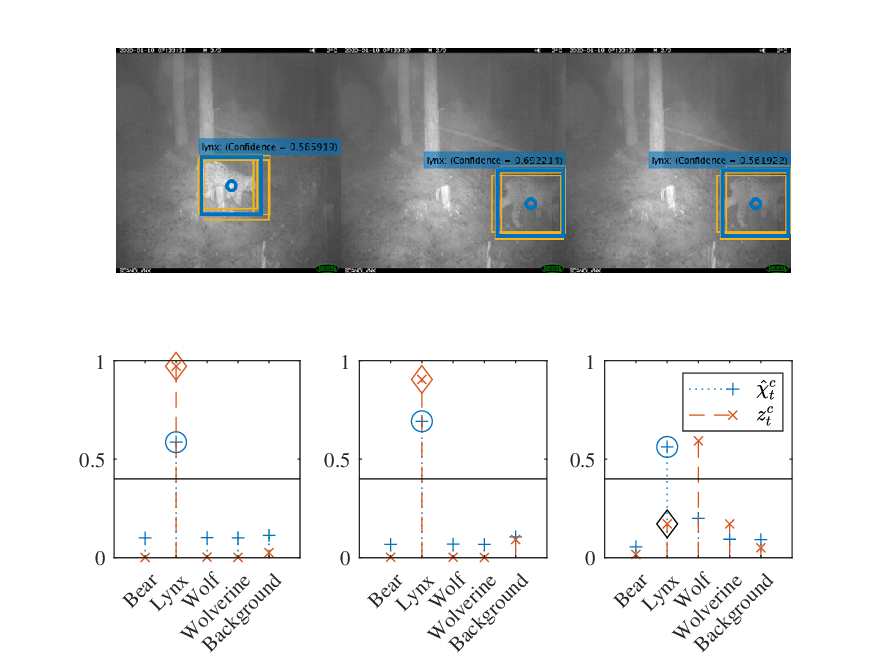}
	\includegraphics[trim={1cm 0cm 0.5cm 6cm},clip,width=0.355\textwidth]{fig/ssd_resnet_lynx_badmeas_2023_10_06}
	\caption{ Robust estimation of the \pmf for the object class in a sequence of images. Top: The camera image sequence of the lynx with the estimated bounding box $\hat{\chi}_t^b$ in thick blue and the measurement from the detection algorithm in thin yellow, $z^{b,(i)}$.   Bottom: In blue, the estimated \pmf denoted $\hat{\chi}^c_t$ from \eqref{eq:estimated_class}, and in orange,  the measurement of the \pmf $z_t^c$ from \eqref{eq:meas}. In black is a decision line on whether a track is lost. The marker in the \pmf plot denotes the class of the estimated track. It changes color when the track is lost.    }
	\label{fig:two_good_one_bad}
\end{figure}

\subsection{Measurement model}
Instead of using \nms and only keeping the most likely detection of an object, the $B$ most likely anchor boxes/proposals could be used. That is 
\begin{align} \label{eq:proposals}
 z_t^{b,(i)} \!  \triangleq \! f^b(x_t;\hat{\theta}_N)^{(i)} \! \!, \text{ }  z_t^{b,(i)} \! \triangleq \! f^c(x_t;\hat{\theta}_N)^{(i)}\!  , \text{ } i \!= \! 1, \ldots, B.
\end{align}
These proposals in \eqref{eq:proposals} are used to create measurements to the tracking algorithm, $z^b_t$ and $z^c_t$. More precisely, a weighted mean of these proposals is used, i.e.,
\begin{subequations}
	\begin{align}
	z^b_t =\sum_{i=1}^{B} w_i z_t^{b,(i)}, \quad  z^c_t = \sum_{i=1}^{B} w_i z_t^{c,(i)}. \label{eq:meas}
	\end{align}
	The weights are chosen proportional to the relative confidence that the proposed bounding boxes include an object of the most likely class, i.e., the weights are given as
	\begin{align}
	\tilde{w}_i =p(y = \hat{y}^{\ast} | z^{c,(i)}) , \quad w_i = \frac{\tilde{w}_i}{\sum_{j=1}^{B} \tilde{w}_j},
	\end{align}
\end{subequations}
where $ \hat{y}^{\ast}$ denotes the most probable class of the object that is in the image, see \eqref{eq:mostprobable}.
If there are multiple objects in the images, an association process using IoU can be used to create multiple measurements per image. However, to make the notation more concise, it will again be assumed that the images only include one object.
Further, the covariance of the measurement can also be computed as
\begin{align}
R^b_t &= \sum_{i=1}^{B} w_i (z_t^{b,(i)}-z^b_t)(z_t^{b,(i)}-z^b_t)^\top,
\end{align}
which is used in the \kf.
Note that this is not commonly done in the literature.%
\begin{table}[tb!]
	\centering
	\caption{ Number of lost tracks at the last image in the sequence.  }
	\begin{tabular}{ l | c  c  }
		\hline
		Detection using & Proposed, $\hat{\chi}_{t}^c$, \eqref{eq:estimated_class} & Standard, $ z^c_t$, \eqref{eq:meas}  \\ [0.5ex]
		\hline
		Number of lost tracks & 2/20&  20/20  \\
		[1ex]
	\end{tabular}
	\label{table:lost_traks}
\end{table}

\section{Wildlife conservation}
With camera technology getting cheaper, more compact, and more durable, it enables the use of edge devices for camera surveillance systems over larger areas. One application where this is useful is monitoring animals in national parks and animal sanctuaries. 
Carnivores such as lynx and wolves are keystone species in the European wilderness \cite{hoeks2020mechanistic}. Hence, there have been attempts to reintroduce them by organizations such as Rewilding Europe. However, collaboration and acceptance from the general public are important to reintroduce them successfully. Here, a camera monitoring system can be used to warn the general public and to count the number of individuals \cite{wwf_camera}.
Apart from monitoring where the animals are, distributed camera systems on edge devices can be used as a warning system for poaching \cite{tyden2020}. 
Camera traps provide a sequence of images, which often might be of bad quality and taken in poor lighting conditions, and of which many do not contain any object of interest. However, it is still important for the ranger monitoring the park to understand what is going on without spending too much time on false positive detection of objects.
There is a limited amount of training images for training the detection algorithms.  
One approach to increase the accuracy with the limited data available is to propagate the information over the entire image sequence.

\section{Experiments}
This paper uses image sequences from camera traps from Swedish forests. The traps belong to a project to monitor the four Swedish top carnivores, i.e., bear, lynx, wolf, and wolverine.
For the first experiment, a sequence of two correct measurements is followed by an incorrect one. The incorrect measurement is a copy of the previous measurement but where $z^c_t$ is artificially changed. Here, 20 such sequences are used to evaluate the method for the filtering problem. A track is considered lost if $\max \hat{\chi}_{t}^c<0.4$ or the most likely non-background class is changed.

The backbone \nn used is a ResNet50 pre-trained on the ImageNet dataset \cite{deng2009imagenet}. The \ssd is used as a detection algorithm and is trained using stochastic gradient descent with momentum. For the first image, $\chi_0^c$ is initialized as a flat distribution where all classes are equally likely, and $\chi_0^b$ is initialized as the object's true position. The implementation of the \ssd is done using the deep learning toolbox in \matlab. 

\begin{figure}[tb!]
	\centering
	\includegraphics[trim={3cm 22cm 3cm 3cm},clip,width=0.42\textwidth]{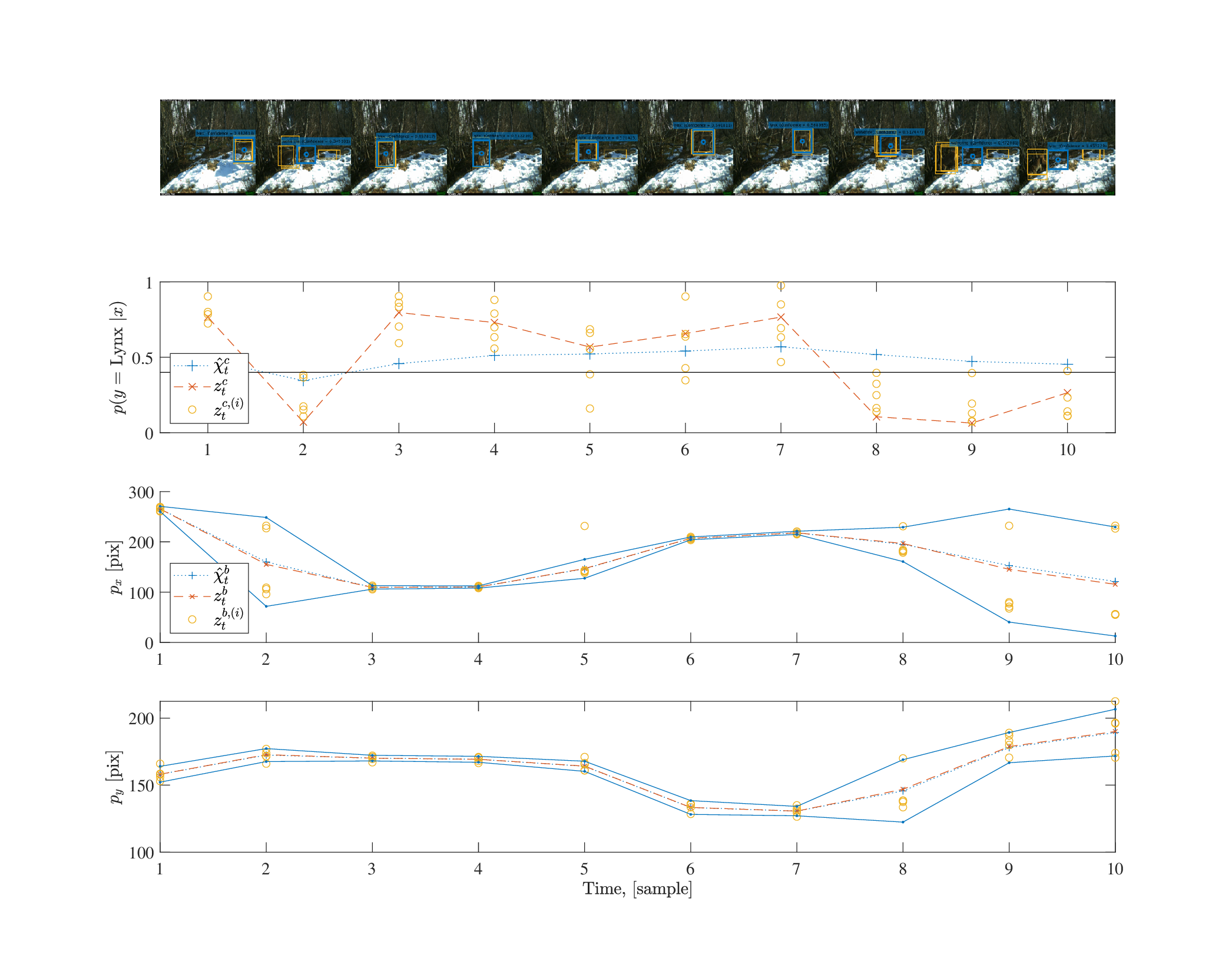}
	\includegraphics[trim={3cm 8cm 3cm 8cm},clip,width=0.420\textwidth]{fig/ssd_lynx_c_2023_10_04}
	\includegraphics[trim={3cm 1cm 3cm 25.3cm},clip,width=0.420\textwidth]{fig/ssd_lynx_c_2023_10_04}
	\caption{ On the top is the image sequence of a lynx, followed by the probability of a lynx in the image and the tracking of the $x$-position of the bounding box $p_x$ over time. The estimated states are shown in blue, the measurement to the filter in orange, and the measurements from the individual anchor boxes in yellow. In black is a decision line on whether a track is lost.     }
	\label{fig:three graphs}
\end{figure}

\cref{fig:two_good_one_bad} show one of the 20 sequences where the measurement of the class for the last frame is artificially changed, but by using the information from previous frames, the track survives. The result for the number of lost tracks for the last images in the sequence for all the sequences can be seen in \cref{table:lost_traks}.
Notice that with a more intricate choice of $K^c_t$, it would have been possible for all tracks to survive. However, with this simple choice, we still get good results.
Without using information from previous frames,  the track of the object is lost for all sequences, e.g., see the diamond marker in \cref{fig:two_good_one_bad}.
In \cref{fig:three graphs}, an experiment is shown where a lynx is tracked over ten frames. It can be seen that using the information from the previous frame results in a more robust prediction, i.e., even though the measurement from the \ssd is incorrect, $\hat{\chi}_t^c$ indicates the correct class. It is also shown how the position of the bounding box is tracked using the specified measurement covariance.

\section{Summary and Conclusions}
This paper proposes a filtering framework for the multi-object tracking problem such that information regarding the object class in previous frames can be used to classify the current frame. The problem can be split into two parts. Firstly, to detect the object using a standard algorithm, and secondly, to specify a state-space model where the output from the detection algorithm is used as measurements.
Since the method is based on standard detection algorithms, it is a stand-alone method that can be used out-of-the-box for any object-detection algorithm that uses proposal anchor boxes. Further, it is shown how to quantify the covariance of the position of the detected object.

The method is evaluated in real-world image sequences from camera traps to monitor carnivores in the Swedish forest. The method is shown to improve the robustness of the prediction. The improved robustness can be seen in experiments where even if the classification from one image in the sequence is incorrect, using information from previous images in the respective sequence can help correct the prediction in 18 of 20 sequences.
%

 \addtolength{\textheight}{-11.5cm}   



%






\bibliography{IEEEabrv,mybibref5G}

\end{document}